# Generating Long-term Continuous Multi-type Generation Profiles using Generative Adversarial Network

Ming Dong, *Senior Member, IEEE*

*Abstract*— Today, the adoption of new technologies has increased power system dynamics significantly. Traditional long-term planning studies that most utility companies perform based on discrete power levels such as peak or average values cannot reflect system dynamics and often fail to accurately predict system reliability deficiencies. As a result, long-term future continuous profiles such as the 8760 hourly profiles are required to enable time-series based long-term planning studies. However, unlike short-term profiles used for operation studies, generating long-term continuous profiles that can reflect both historical time-varying characteristics and future expected power magnitude is very challenging. Current methods such as average profiling have major drawbacks. To solve this challenge, this paper proposes a completely novel approach to generate such profiles for multiple generation types. A multi-level profile synthesis process is proposed to capture time-varying characteristics at different time levels. The proposed approach was evaluated based on a public dataset and demonstrated great performance and application value for generating long-term continuous multi-type generation profiles.

*Index Terms*—Power System Planning, Generation Forecast, Machine Learning

## I. Introduction

Traditonally, utility companies have commonly used representative discrete values obtained from long-term forecast such as peak, average or percentile values as input for long-term system planning studies. Today, driven by new technologies such as renewable energy, energy storage and demand response, system dynamics has increased dramatically [1]. As a result, system planning now requires much more granular studies based on time-series data instead of discrete values. For example, 8760 hourly profile is becoming a new requirement for planning studies in today's power industry [1-6]. This level of data resolution can adequately model the time-varying behaviours of loads and generations and the combined effect to a power system. Based on such profiles, accurate time-series based planning studies and simulations can be performed to realistically identify future system deficiencies and derive proper planning decisions. However, comparing to only a few values, creating 8760 generation profiles is much more challenging as the historical time-varying characteristics are difficult to capture and summarize as hard rules or assumptions. Currently, for creating continuous generation profiles, the industry has made the following attempts [7]:

- Selecting one or making an average profile from a few historical yearly profiles and apply it to future years. This method is straightforward but has a few major drawbacks: using one historical profile may not be representative and other historical data is ignored; averaging profiles can cause the loss of dynamic details that are important for time-series analysis; using one historical or the averaged profile can cause diversity issues. Profiles of different years and/or generation sites will look the same and are unrealistic; this method is difficult to incorporate expected power magnitude that forecasting engineers can predict based on their analysis of area load, new generation capacity, production cost, renewable energy policy, long-term weather senarios and etc. [6]
- Randomly sampling daily profiles. This method randomly picks a daily profile from an existing profile database and scales it to match with the expected power magnitude. This method may be applicable to some operation studies when only separate daily profiles are required but is not applicable to long-term continuous profiles. The is because multiple daily profiles cannot be direcly connected. No data continuity will exist between two adjacent days unless the starting and ending point values of each day are manually altered for smooth transition, which is difficult to do and also can damage the inherent profile patterns. Furthermore, with limited existing daily profiles, this method also lacks diversity, especially when profiles of multiple future years are often required for long-term planning studies.
- Simulating generation behaviours for every future hour. The method is difficult to apply because the simulation models can be quite complex and costly to build and often requires special commercial software or service[8]. The accuracy relies on a large set of generation and external assumptions that are often difficult to set. The results are sensitive to assumptions and can vary significantly between cases. In reality, the simulation is often unable to reproduce historical data characteristics. Also, the simulation process can be computationally expensive and may need numerous trials to land on convincing results.

To address the above challenges, this paper proposes a data-driven approach based on Generative Adversarial Network (GAN) models to automatically capture the historical time-varying characteristics of different types of generations and more importantly, reproduce such characteristics when generating future time-series profiles. In recent years, GAN has become a hot topic in the AI research and application field. It has been applied successfully to capture image characteristics and create very similar images that human cannot tell apart from the real ones. The idea of synthesizing time-series generation profiles using GAN is inspired by such accomplishments [9-11].

The current research on applying GAN to solve power system problems is still at an early stage: [12] uses GAN to generate missing PMU data for power system dynamic security assessment; [13-15] uses GAN to generate building and appliance loading data to augment limited training datasets for relevant machine learning tasks; [16] uses GAN to increase solar irradiance data to enhance weather classification accuracy. As can be seen, [13-16] essentially tried to solve the small sample learning problem by using GAN; [17] uses GAN to extract characteristics from smart metering data and replaces real smart metering data with synthesized smart metering data so that the privacy of real smart metering data can be protected; [18] uses GAN to add load forecast uncertainty to the single-point deterministic forecast results and hence make the load forecast result probabilistic. Compared to the above research works, the research problem we discuss in this paper is unique and is focused on producing long-term continuous hourly profiles for multiple generation types with specific requirements summarized as below:

- The generated profiles should be able to reflect the historical time-varying characteristics;
- The generated profiles should be able to reflect the expected generation levels obtained from long-term generation forecast;
- The generated profile should be long-term and continuous with an hourly resolution(8760 points a year);
- The proposed method should be applicable to different generation types;
- The proposed method should be able to produce unlimited number of different profiles to meet the diversity requirement.

This paper is organized as follows: Section I explains the research background, long-term continuous generation profile requirements and presents literature review on relevant research works; Section II proposes the multi-level profile synthesis process which is a novel and critical process for creating continuous long-term profiles by using GAN; Section III proposes two GAN systems with different structures to serve the purpose of generating profiles for multiple generation types; Section IV focuses on the explanation of the training process for GAN models and proposes a new loss function for Multi-type GAN; Section V further proposes evaluation metrics and provides detailed validation of the proposed method on a public long-term generation dataset with comparison to the traditional average profiling method. It also briefly discusses a Monte-Carlo based post-processing step in order to add random generation outages to the profiles. The evaluation demonstrates great performance and application value of the proposed method. In the end, some future improvement opportunities are also identified and discussed

## II. Conclusions and Discussions

This paper proposed a completely novel approach for generating long-term continuous multi-type generation profiles using GAN models. To achieve this goal, this paper firstly analyzes the drawbacks of traditional methods and summarizes the requirements. The novelties of this paper can be summarized as below:

- it proposes a unique multi-level profile synthesis process with the "starting-point" trick and "duty-cycle" trick that is able to capture time-varying characteristics at different time levels and reflect the expected power magnitude;
- it proposes a new Conditional GAN structure called Multi-type GAN and its unique loss function
- it proposes a series of metrics for evaluating the quality of the GAN generated profiles.

The approach was applied to a public dataset. Different variations were implemented and compared from different requirement perspectives. The result shows that the proposed approach is able to:

- generate realistic profiles that capture time-varying characteristics at different time levels for different types of generations;
- generate profiles with specific magnitude that complies with long-term forecasting expectation;
- and generate diverse profiles that are never repeated.

The proposed approach is an effective solution for generating long-term continuous generation profiles for multiple types of generations in order to serve the increasing need of dynamic long-term system assessment and simulations. Based on the existing effort and outcome, there is still improvement room in the future:

- More advanced Conditional GAN systems can be investigated such as using a third neural network for condition match judgment and using special embedding layers to incorporate condition information [26];
- Generate profiles with more granular condition information such as specific operation characteristics in addition to general type. For this purpose, a Style-GAN based solution may need to be developed [27].